\documentclass[10pt,twocolumn,letterpaper]{article}

\usepackage{cvpr}
\usepackage{times}
\usepackage{epsfig}
\usepackage{graphicx}
\usepackage{amsmath}
\usepackage{amssymb}
\usepackage{comment}
\usepackage{bbm}
\usepackage{pdfpages}
\usepackage{lipsum}

\newcommand\blfootnote[1]{%
  \begingroup
  \renewcommand\thefootnote{}\footnote{#1}%
  \addtocounter{footnote}{-1}%
  \endgroup
}

\newcommand{\Linchao}[1]{\textcolor{red}{[L]}}
\newcommand\given[1][]{\:#1\vert\:}

\usepackage[pagebackref=true,breaklinks=true,letterpaper=true,colorlinks,bookmarks=false]{hyperref}

\cvprfinalcopy % *** Uncomment this line for the final submission

\def\cvprPaperID{6570} % *** Enter the CVPR Paper ID here

\ifcvprfinal
\fi
\begin{document}

%%%%%%%%% TITLE
\title{Sim-Real Joint Reinforcement Transfer for 3D Indoor Navigation}
%\title{Sim-Real Joint Reinforcement Transfer for 3D Indoor Navigation}

%\author{Fengda Zhu\and Linchao Zhu \and Yi Yang \\
%CAI, University of Technology Sydney\\
%\\
%{\tt\small \{zhufengdaaa, zhulinchao7, yee.i.yang\}@gmail.com}
%}

%\author{Fengda Zhu \hspace{2em} Linchao Zhu \hspace{2em} Yi Yang \\
%CAI, University of Technology Sydney\\
%{\tt\small \{zhufengdaaa, zhulinchao7, yee.i.yang\}@gmail.com}
%}

\author{Fengda Zhu$^\dag$\hspace{6mm}
Linchao Zhu$^\S$\hspace{6mm} 
Yi Yang$^\S{}^\ddag$ \\
%
%\newline
%\,Corresponding author: Yi Yang.
%} \\  %\thanks{Corresponding author.}  \\
$^\dag$UTS-SUSTech Joint Research Centre, Southern University of Science and Technology\\
$^\S$CAI, University of Technology Sydney \hspace{2mm}  $^\ddag$Baidu Research   \\
{\tt\small \{zhufengdaaa,zhulinchao7\}@gmail.com \hspace{2mm} Yi.Yang@uts.edu.au}}

% For a paper whose authors are all at the same institution,
% omit the following lines up until the closing ``}''.
% Additional authors and addresses can be added with ``\and'',
% just like the second author.
% To save space, use either the email address or home page, not both

\maketitle
%\thispagestyle{empty}

%%%%%%%%% ABSTRACT
\begin{abstract}
%To train agents for navigation, many simulated environments for synthetic or real scenes have been proposed. However, it is difficult to train agents in a real 3D environment since current real environment has much less scenes. 

% There has been an increasing interest in 3D indoor navigation, where a robot in an environment move to a target according to an instruction. It is quite labour intensive to obtain sufficient real environment data for training the robot. In this paper we propose to use synthetic environment to r. 

There has been an increasing interest in 3D indoor navigation, where a robot in an environment moves to a target according to an instruction. To deploy a robot for navigation in the physical world, lots of training data is required to learn an effective policy. It is quite labour intensive to obtain sufficient real environment data for robots training while synthetic data is much easier to construct by rendering. 
Though it is promising to utilize the synthetic environments to facilitate navigation training in the real world, real environment are heterogeneous from synthetic environment in two aspects. First, the visual representations of the two environments have significant variances. Second, the house plans of the two environments are rather different.  
Therefore, two types of information, \ie, visual representation and policy behavior, need to be adapted in the reinforcement model.
The learning procedure of visual representation and that of policy behavior are presumably reciprocal. 
%In this paper we propose to use synthetic environment to help training model in real environment. 
We propose to jointly adapt visual representation and policy behavior to leverage the mutual impacts of environment and policy. 
Specifically, our method employs an 
adversarial feature adaptation model for visual representation transfer and a policy mimic strategy for policy behavior imitation.
The experimental results show that our method outperforms the baseline by 21.73\% without any additional human annotations.

%Thus transfer learning approaches have been proposed, motivated by improving real environmental training by utilizing synthetic data. We propose Joint Reinforcement Transfer Network(JRTN), our reinforcement model transfer method consisting of two methods: Adversarial Feature Adaptation and Policy Mimic. As a result, our final model outperforms the baseline of only real environment training data by 19.47\%. 

\end{abstract}

\begin{figure}[t]
\centering
\includegraphics[width=0.98\linewidth]{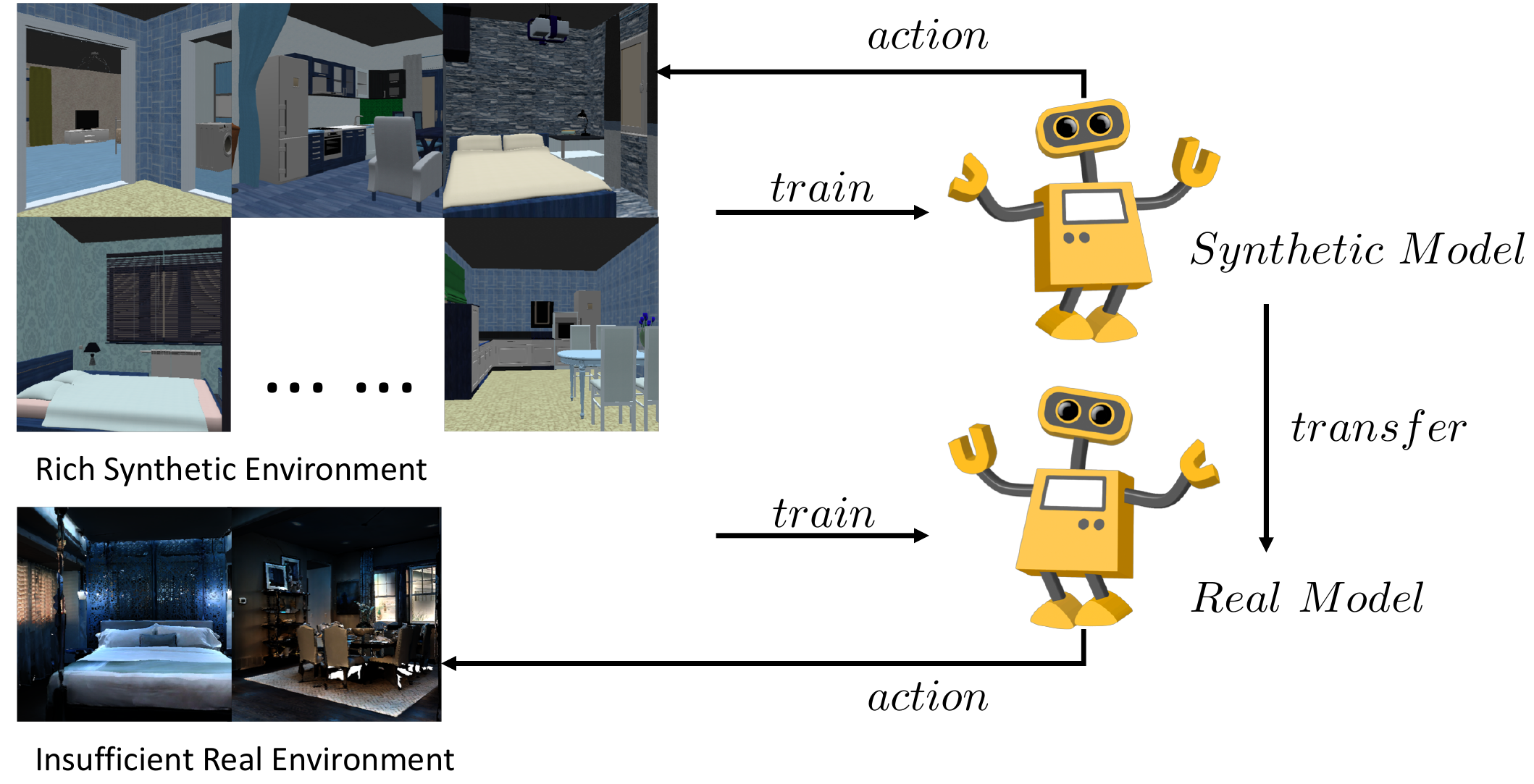}
    \caption{
This figure illustrates the general framework about transferring knowledge from synthetic model (top) to real model (down). In testing, the two models receive images and predict actions (turn left, turn right or forward) to navigate in environments. 
%Data scale is much larger in synthetic environment because of a great number of house structures and more kinds of customized indoor texture. 
    }
\label{fig:demo}
\end{figure}

%%%%%%%%% BODY TEXT
\section{Introduction}
\blfootnote{Part of this work was done when Yi Yang was visiting Baidu Research during his Professional Experience Program.}
\blfootnote{Corresponding author: Yi Yang.}

Autonomous indoor navigation is a problem in which the robot navigates to a target according to an instruction within a building such as house, office and yard. 
This task will benefit many applications where a robot takes over human being's job, such as house cleaning, package delivery and patrolling.  
Solving indoor navigation in a 3D environment is the basis of these mobile robotic applications in the real-world scenario. 

Earlier works adopt imitation learning methods including behavior cloning~\cite{argall2009survey} and DAGGER~\cite{ross2011reduction} for 3D navigation tasks. These approaches train a robot to emulate an expert, such as the shortest path in indoor navigation. These approaches fail into over-optimization, which suppress a larger set of close-optimal solution. Deep reinforcement learning approaches based on actor-critic, \eg, A3C~\cite{mnih2016asynchronous} and UNREAL~\cite{jaderberg2016reinforcement} are widely used in recent researches~\cite{mousavian2018visual, yu2018guided, wu2018building, savva2017minos}. Advantage of these end-to-end approaches is that it discretizes the agent and state space~\cite{zhu2017target} and explores explicit map representations for planning~\cite{mirowski2016learning}. 

There has been impressive progress on deep reinforcement learning (RL) for many tasks, such as Atari video games, GO~\cite{mnih2013playing, silver2017mastering}, robot control~\cite{duan2016benchmarking}, self-driving~\cite{sallab2017deep} and navigation~\cite{wu2018building}. 
However, it is difficult to apply reinforcement learning to the physical environment since sampling a large number of episodes for training a robot is time-consuming and even impossible. Thus, recent researches focus on the RL model to learn policy in a simulated environment rather than in the physical world to resolve the problem~\cite{wu2018building, savva2017minos}.  
There are two types of environments that can be used to train a robot. The first one is rendered synthetic environment.  A represent work is SUNCG~\cite{song2017semantic}, which produces a 3D voxel representation from a single-view depth map observation. The second one is reconstructed environment such as  Matterport3D~\cite{chang2017matterport3d}, which consists of real images captured with a Matterport camera.

One problem of only using the Matterport3D dataset for training, however, is the lack of diversity of scenes. the SUNCG dataset~\cite{song2017semantic} consists of over 45,622 synthetic indoor 3D scenes with customizable layout and texture, while Matterport3D~\cite{chang2017matterport3d} contains only 90 houses. Lack of training houses and scenes can significantly hamper performance because of overfitting. Models trained on the SUNCG dataset, on the contrary, is more capable of generalizing to unseen scenes. 

In this paper we propose a joint framework, namely Joint Reinforcement Transfer (JRT), to resolve the problem of lacking training data in the real environment. 
Our premise is that synthetic training data is much easier and cheaper to obtain than real data. 
As shown in Figure~\ref{fig:demo}, our algorithm integrates adversarial feature adaption and policy mimic into a joint framework. 
The joint learning of the adversarial feature adaption and policy mimic makes them mutually beneficial and reciprocal. 
In this way, the feature adaption and policy mimic are tightly correlated.
The adversarial feature adaptation not only generates better representation well fits real environment, but is also more suitable for policy imitation. 

%Thus we have created a discriminative semantic analysis framework based on a tightly coupled intermediate representation.

 %We demonstrate our idea on reinforce models using the actor-critic approaches~\cite{mnih2016asynchronous, jaderberg2016reinforcement}. This reinforce model can be divided into two parts: visual mapping function and policy function. The visual mapping function convert visual information(RGB image in this paper) to visual embedding, which is a vector representing visual semantic feature. Policy function predicts action distribution based on visual semantic feature. We apply feature adaption and policy transfer on visual mapping function and policy function respectively. 

It is intuitive to transfer visual feature from synthetic domain to real domain, so that we can directly adopt knowledge learned from synthetic environment without changing the policy function. This kind of method, also called Domain Adaptation, is widely applied in numerous tasks~\cite{long2013transfer, saenko2010adapting, yang2007cross,Yawei2019Taking, duan2009domain, tzeng2017adversarial}. Inspired by the idea of Adversarial Discriminative Domain Adaptation, we use the adversarial loss to supervise our transfer training process. Compared to image translation methods like CycleGan~\cite{zhu2017unpaired}, which directly translates the visual image with GAN~\cite{goodfellow2014generative,NIPS2017_7159, pmlr-v80-cao18a, zhu2019dynamic}, our approach drops the redundant steps that generates target images and extracting feature for target domain. Therefore, our model is pimplier in framework and less parameter in model so that be easy to converge. 

\begin{figure}[t]
\centering
\includegraphics[width=0.98\linewidth]{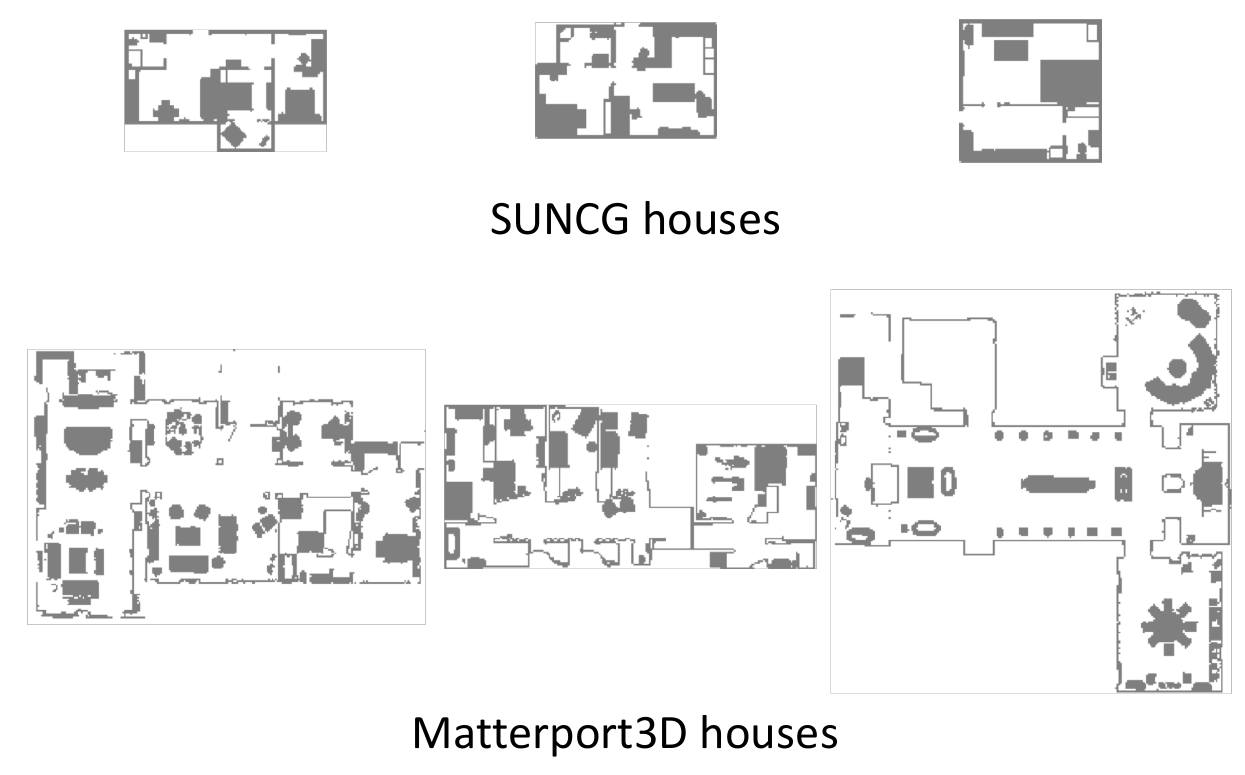}
    \caption{
This figure shows the different house plans of the SUNCG dataset and Matterport3D dataset. Houses in SUNCG are smaller and simpler in layout. Meanwhile, houses in Matterport3D have more rooms and are more complicated in house plan. 
%Data scale is much larger in synthetic environment because of a great number of house structures and more kinds of customized indoor texture. 
    }
\label{fig:houses}
\end{figure}

In addition, the policy could be tuned under the supervision of teacher trained on synthetic environment. 
Our motivation is two-fold. First, the layout of real environment is not exactly the same as the synthetic environment. As shown in Figure \ref{fig:houses}, houses of real environment has more rooms and much more connections between rooms, which means more complicated structure. Second, some knowledge learned in synthetic scene, such as finding doors and bypassing obstacles is also helpful in real environment navigation. Thus, we propose our Joint Reinforcement Transfer(JRT) to integrate the two training stages together. 

Our experimental results show that this joint method outperforms baseline with finetune by 21.73\% without any additional human annotations.
Also, By qualitatively visualizing the results of the two adaptation methods, we shows the two parts are mutually enhanced to benefit real-world navigation.

%It shows find both of them have unique but important effect on model performance. 

% To summarize our contributions: We propose novel approach to augment reinforce model learning in real environment by synthetic data. We believe that knowledge model learned from synthetic data can be transferred from two aspects: feature representation and policy behavior. Our approach is divided into two joint methods: Adversarial Feature Adaptation and Policy Mimic. 

\section{Related Work}
The proposed work is related to transferring reinforce policy trained on virtual environment to real environment. In this section, we briefly review several methods on 3D Navigation, Indoor Environment, Domain Adaptation and Policy Transfer with high relation to our topic.

%\noindent\textbf{3D Navigation}
%Earlier works adopt imitation learning method including behavior cloning\cite{argall2009survey} and DAGGER\cite{ross2011reduction} to teach agents. In such approaches, agent is trained to emulate an expert, usually the shortest path in indoor navigation. These approaches fail into over-optimization, which suppress a larger set of close-optimal solution. Deep Reinforcement learning approaches based on actor-critic and its variances A3C\cite{mnih2016asynchronous} and UNREAL \cite{jaderberg2016reinforcement} are widely used in recent researches\cite{mousavian2018visual, yu2018guided, wu2018building, savva2017minos}. One advantage of this intrinsic, end-to-end approach is that it discretizes the agent and state space \cite{zhu2017target} and explores explicit map representations for planning \cite{mirowski2016learning}. 

\noindent\textbf{3D Indoor Environment}
There has been a rising interest in indoor reinforce environment. House3D~\cite{wu2018building} is a manually created large scale environment. AI2-THOR~\cite{kolve2017ai2} is an interactable synthetic indoor environment. Agent can interactive with some interactable objects such as open a drawer or pick up a statue. CHALET~\cite{yan2018chalet} is another interactable synthetic indoor environment with larger interactive action space. Recent works tend to focus on simulated environment based on real imagery. However, the scale of these datasets are quite small compared with virtual datasets. Active Vision dataset~\cite{yan2018chalet} consists of dense scans of 16 different houses. And Matterport3D ~\cite{chang2017matterport3d} is a larger, multi-layer environment. Minos is a cross domain environment which consist both House3D and Matterport3D. MINOS provides similar setting of both environments, which is convenient for our environmental transferring experiment. 

\noindent\textbf{Domain Adaptation}
The problem of domain adaptation has been widely studied and arises in different visual application scenarios such as image classification~\cite{long2013transfer, saenko2010adapting}, object detection~\cite{yang2007cross} and action recognition~\cite{faraji2011domain}. Prior work such as~\cite{huang2007correcting, gretton2009covariate} used Maximum Mean Discrepancy(MMD)~\cite{quinonero2008covariate} loss to minimize the difference between the source and target feature distributions. MMD computes the difference of data distributions in the two domains. Parameter adaptation, another kind of early method used in~\cite{yang2007cross, duan2009domain, he2019filter}, adapt the classifier like SVM trained on the source domain. 

In contrast, Recent works introduce adversarial approach into domain adaptation. Generative Adversarial Network(GAN) ~\cite{goodfellow2014generative}, lets generator G and discriminator D compete against each other, is widely applied because of its powerful ability for learning and generalizing from data distribution.  
Gradient Reversal~\cite{ganin2016domain} directly optimizes the mapping by reversing the gradient of discriminator. 
To go one step further, Adversarial Discriminative Domain Adaptation~\cite{tzeng2017adversarial} uses a target generative model optimized by adversarial loss to learn source feature distribution. This method can better model the difference in low level features. There are several work~\cite{handa2016understanding, xu2014learning} focus on Synthetic-to-Real visual image translation and a benchmark called VisDA~\cite{peng2017visda} has been proposed recently. 

\noindent\textbf{Policy Transfer}
Some works have explicitly studied action policy transferring in reinforcement learning senario in different way. 
Hinton \etal ~\cite{hinton2015distilling} proposes the method of Network Distillation, using student model to learn the output distribution of teacher model. 
Policy Distillation~\cite{rusu2015policy} employs this method to transfer knowledge between two environment based on DQN algorithm. Target-driven Visual Navigation~\cite{zhu2017target} is a model for better generalize to new goals rather than new environmental scene. Semantic Target Driven Navigation~\cite{mousavian2018visual} learns policy only based on detection and segmentation result, so that model can be easily transferred to unseen environments. 

\section{Model}

\subsection{Baseline Setup}

We demonstrate our approach of reinforce model transfer based on room goal navigation task. 
Before our approach, we will first present our general model under the framework of standard reinforcement setting. 

According to partially observable Markov decision process (POMDP), we can formulate our problem as (S, A, O, P, R).
The agent starts from a initial state $s\in S$, which is the pose of the agent, a position direction pair. The action $a \in A$ is a discrete set of predefined actions. Observation space is the union of multi-modal sensory input space such as RGB image, depth image and force $O=\left \{ O_{rgb}, O_{depth}, O_{force}... \right \}$. Probabilistic state-action transition function is represented as $p(s_{i+1} \given s_{i}, a_{i})$. Reward $R(s,a)$ is a function related to the Euclidean distance to the goal and normalized time~\cite{savva2017minos}. 

\noindent\textbf{LSTM A3C}
LSTM A3C is a general model of asynchronous advantage actor-critic algorithm. Compared to Feedforward A3C whose policy only based on the current observation, LSTM A3C introduce temporal experience to achieve better performance. 

LSTM A3C starts with a convolution network used as visual embedding module
$$f = CNN(O)$$
$f$ stands for visual feature of current observation. The visual feature sequence from the step 0 to current step $t$ is written as $f_{seq} = \left\{ f_{0}, f_{1}, ..., f_{t}\right\}$. 

Also we have a goal $g$ to indicate which room agent is required to go. $g$ is encoded into a semantic feature vector as $g_{emb}$ by a embedding layer: 
$$g_{emb}=G_{emb}(g)$$

Then visual feature of each step and goal embedding are concatenated to fed to (Long-Short Term Memory) LSTM units for temporal encoding
$$h_{t} = LSTM([f_{t}, g_{emb}], h_{t-1})$$
where $f_{t}$ is the visual feature of current step and $h_{t}$ is the LSTM output vector, encoding the historical information from step $0$ to $t$. Note that we simplify the propagation of memory cells for convenience. 

Follow the architecture of actor-critic,  we have a policy module $a=\pi(a|h_{t})$ to predict the distribution of ongoing actions and value $v=V(h_{t})$ to predict the value of this state. Loss functions of A3C baseline is defined as
$$L_{A3C}=-log\pi(a|h_{t};\theta)(R_{t} - V(h_{t};\theta)) - H(\pi(a|h_{t};\theta))$$

where term $H(\pi(s_{t};\theta))$ represents the entropy of policy. We maximize this term to encourage exploration. 

\noindent\textbf{UNREAL} UNREAL is an advanced version of A3C with several unsupervised auxiliary tasks providing wider training signals without additional training data. This improvement is general so that it can augment many reinforce tasks including navigation in this paper. We will prove that our approach can also be applied to UNREAL model in experiment and achieve the state-of-the-art on our setting. 

\noindent\textbf{Re-Formulation}
In the following sections, we will propose two method, Feature Adaptation and Policy Transfer. 
For simplify notation, we re-define the model as two parts, which consist of a mapping function $M$ and a policy function $P$. 

Mapping function $M$ is fed with observation(only RGB images in our experiment) $x \sim X$ and output visual embedding $f \sim F$. We define $X$ as RGB image distribution and $F$ as feature distribution. $X_{s}, X_{r} \subseteq X$ represent image distribution of synthetic domain and real domain respectively. Similarly, we use  $F_{s}, F_{r} \subseteq F$
 to represent synthetic visual feature and real visual feature.

Policy function $P$ takes $f$ as input and predict discrete action with $softmax$ activation. Our representation of reinforce model is written as
$$\pi(a|s;\theta) = softmax (P(M(o; \theta_{M}); \theta_{P}))$$

We use $L_{policy}$ to represent the united form of actor-critic loss for both A3C and UNREAL. 

% A figure for the pipeline
% A figure for the task
%

\begin{figure*}[t]
\centering
\includegraphics[width=0.98\linewidth]{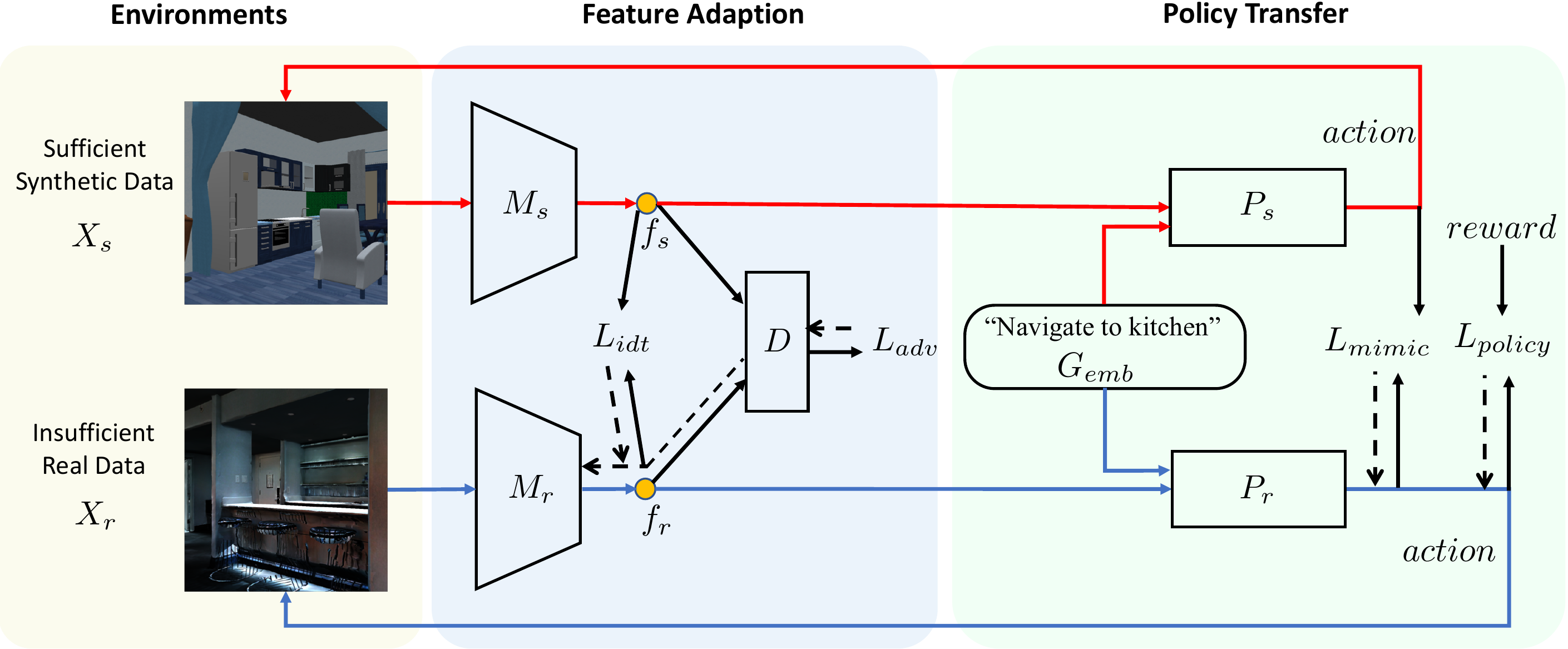}
    \caption{
An overview of our Joint Reinforcement Transfer (JRT). Our model contains two RL models: synthetic model with $M_{s}$ and $P_{s}$ and real model with $M_{r}$ and $P_{r}$.  We use red line and blue line to represent the testing procedure of the two models. Black arrows stand for forward and dotted lines stand for backward in the training procedure.  $f_{s}$ and $f_{r}$ are feature embeddings of synthetic image $X_{s}$ and real image $X_{r}$, respectively. For feature adaptation, we apply adversarial loss $L_{adv}$ on top of discriminator $D$. Additionally, we have identity loss $L_idt$ to regularize the learned semantic embedding. In policy transfer, we have the mimic loss $L_{mimic}$ trained with policy loss $L_{policy}$. 
    }
\label{fig:feature_adaption}
\end{figure*}

\subsection{Adversarial Feature Adaptation}\label{adaptation}

For adapting model trained from simulated environment to real environment, one intuitive idea is mapping visual observation to a latent space with same distribution. Thus the following policy network can be easily applied to real environment since its input distribution will not change significantly. 

Motivated by this, we consider adversarial learning method to adapt the mapping function for real images. Note that we use A3C to demonstrate our approach for simplicity. Following the notation of~\cite{tzeng2017adversarial}, we assume $X_{s}$ as image drawn from a distribution of synthetic image domain $p_{syn}(x,y)$ and $Y_{s}$ as its label. Similarly, we have real image $X_{r}$ and label $Y_{r}$ from $p_{real}(x,y)$. We mark the mapping function for synthetic images as $M_{s}$. Thus we have 

$$f_{s} = M_{s}(x_{s})$$
Here $f_{s}$ also follows a synthetic feature distribution represented as $f_{s} \sim F_{s}$. Our goal is to train a real mapping function $M_{r}$ for visual embedding $f_{r} \sim F_{r}$, where $f_{s}$ and $f_{r}$ belongs to the same distribution. 
$$f{r} = M_{r}(s_{r})$$
$$f_{s}, f_{r} \in S, F{r} = F{s}$$

To build our adversarial learning procedure, we need a binary discriminator $D$ to distinguish whether a feature $f$ is mapped from $X_{s}$ or $X_{r}$. Label $l_{k}$ equal to 1 is when $f$ is mapped from $X_{s}$ and equal to 0 otherwise. Thus the classifier $D$ is optimized by cross entropy loss below

\begin{align}
\nonumber
L_{cls}(X_{s}, X_{r}) = &- \mathbb{E}_{x_{s} \sim X_{s}} \log(D(M_{s}(x_{s}))) \\
\nonumber
&- \mathbb{E}_{x_{r}\sim X_{r}} \log(1-D(M_{r}(x_{r})))
\end{align}

We design our adversarial loss to optimize $M_{r}$ so that $M_{r}$ and $C$ can compete each other

$$L_{adv} = \mathbb{E}_{x_{r}\sim X_{r}} \log(D(M_{r}(x_{r})))$$

The optimization of $M_{r}$, which equal to maximizing $\mathbb{E}_{x_{r}\sim X_{r}}log(D(M_{r}(x_{r})))$, will force the distribution of $F_{r}$ getting closer to $F_{s}$, so that discriminator $D$ is more difficult to distinguish $F_{s}$ and $F_{r}$.  By alternatively optimizing $M_{r}$ and $D$, The visual embedding $f_{s}$ and $f_{r}$ will finally belong to same distribution.

However, $M_{s}(x_{s})$ and $M_{r}(x_{r})$ can still be mapped to different semantic vector in latent space. It is due to adversarial method above does not ensure $f_{s}$ to have the same semantic representation as $f_{r}$. Inspired by the technique of~\cite{taigman2016unsupervised}, we used identity mapping loss to regularize the mapping function $M_{r}$ to be an approximate function of identity mapping when input is a synthetic image $x_{s}$. We use L2 loss because our regularization is applied on feature space. This approach suppose that mapping function $M_{r}$ has good generalization ability and synthetic image shares exact the same semantic space as real image. Thus we propose our identity mapping loss:

$$L_{idt} = \left \| M_{s}(X_{s}) - M_{r}(X_{s}) \right \|_{2}$$

By introducing L2 weight norm for both $M_{r}$ and $D$ respectively, our full objective of feature adaption is written as:
$$L_{total} = L_{adv}+L_{cls}+ \lambda_{1} L_{idt}+\lambda_{2} L_{norm}$$

Although $L_{adv}$ and $L_{cls}$ are in same magnitude, we need to balance regularization terms by $\lambda_{1}$ and $\lambda_{2}$. How performance influenced by loss weight are fully discussed in experiment section. 
%\subsubsection{Single Frame}

We use unpaired images from synthetic environment and real environment to train our model. 
As shown in Figure~\ref{fig:feature_adaption}, our gradient are only computed from adversarial loss $L_{adv}$ and identity loss $L_{idt}$. Instead of reinforcement training procedure, we train model following the framework of ADDA~\cite{tzeng2017adversarial}. Parameter of synthetic mapping function $M_{s}$ are fixed while $M_{r}$ and discriminator $D$ is updated per step. We sample batches of images every training step, compute gradient from two losses and then update $M_{r}$ and $D$. 

In testing, $M_{r}$ is connected with policy function $P_{r}$. Without Policy Transfer method, we can just simply copy the parameter from $P_{s}$ to $P_{r}$ and achieve a significant performance improvement from the adaptation of $M_{r}$. 

\subsection{Policy Mimic}\label{mimic}
Previously we have a powerful original policy network trained on large scale synthetic data and a mapping function to adapt visual embedding to real environment. 
However, due to the huge gap between synthetic environment and real environment, adapting visual embedding only is not enough for reinforce model transfer. 

Therefore we introduce an auxiliary approach to transfer our reinforce policy. 
We find it is necessary to combine transferred policy with knowledge gained from real environment. 
Since our task is to predict next action by classification, we introduce Policy Distillation method~\cite{rusu2015policy}, to transfer knowledge leaned from synthetic environment. 

Our baseline trained on synthetic environment acts as a teacher model $T$. A student model learns knowledge from $T$ is written as $S$. Both model is fed by real environment data. And the action probability $p$ predicted by $T$ is served as soft label to train $S$, which is also called mimic. Student model $S$ is trained with a log-likelihood loss to predict the same action distribution: 

$$p = softmax(P_{s}(M_{s}(x_{r})))$$

\begin{align}
\nonumber
L_{mimic} = &- \mathbb{E}_{x_{r} \sim X_{r}} [ p \cdot log(softmax(P_{r}(f_{r})))) \\
\nonumber
&- \mathbb (1-p) \cdot (1-log(softmax(P_{r}(f_{r}))))]
\end{align}

Different from~\cite{rusu2015policy}, which student model only supervised by $a_{best}=argmax(q)$ our student model $S$ learns the full distribution of $p$ from teacher $T$. This loss function enables student learn more knowledge such radios of very small probabilities from soft labels, as demonstrated in~\cite{hinton2015distilling}. 

We optimize $L_{mimic}$ together with $L_{policy}$ in real environment. By balancing the weights between reinforce loss and policy transfer loss, our full objective of Policy Transfer is defined as 

$$L_{total} = L_{policy}+\lambda L_{mimic}$$

Note that loss weight $\lambda$ can be different when $L_{mimic}$ is trained together with UNREAL~\cite{jaderberg2016reinforcement} since UNREAL have several auxiliary tasks which make the magnitude of $L_{policy}$ slightly larger. 

Unlike Feature Adaptation, we follow the reinforcement learning framework of~\cite{mnih2016asynchronous}, training model in real environment only. As shown in Figure~\ref{fig:feature_adaption}, synthetic model and real model take the same input from real environment $X_{r}$. $M_{r}$ is pretrained by adaptation method last section. Parameters of $M_{s}$, $P_{s}$ and $M_{r}$ is fixed and only $P_{r}$ is tuned by mimic loss $L_{mimic}$ and policy loss $L_{policy}$. 

In testing, as the blue line shown in Figure~\ref{fig:feature_adaption}, $P_{r}$ is connected at the bottom of $M_{r}$. Tuning policy function $P_{r}$
 can get another significant performance improvement.

\section{Experiments}
\subsection{Datasets and Baseline Setup}

Our setting is mainly based on MINOS dataset~\cite{savva2017minos}, for MINOS provide unified interfaces for both synthetic and real environments. Same configuration of scenes, multi-modal sensory inputs and action space facilitate implementation. 

% We build our baseline models simply trained on each data environment with different reinforce algorithms. However, following our goal, we test all these models on Matterport3D test split. 

\noindent\textbf{Synthetic Environment}
SUNCG~\cite{song2017semantic} is a large scale environment consists of over 45,622 synthetic indoor 3D scenes. It renders image observation with 3D occupancy and semantic labels for a scene from a single depth map. Following the setting of MINOS,  we manually select a subset of 360 single-floor houses fit for room navigation. A large range of houses in SUNCG, however, are not appropriate for our task. For some scenes, rooms are not connected with each other and other scenes are not even houses. SUNCG scenes are split into 207/76/77 for training, validation and testing respectively. 

\noindent\textbf{Real Environment}
Matterport3D~\cite{chang2017matterport3d} is a multi-layer real-like environment with 90 scenes. We follow the same splitting strategy as original dataset. It has 61 rooms for training and 18 rooms for testing. We sample 10 episode for each test house for total 180 episode as our test set. 

% During training, scenes is randomly sampled. We sample episodes randomly in houses while length is restricted. All models are tested on the test episodes sampled before. We keep the distribution of episodes' length for testing as same as training. 

\noindent\textbf{Baselines}
We consider two algorithms, A3C and UNREAL, which trained on SUNCG and Matterport3D from scratch as our baseline models. For each baseline model, we adopt RMSProp solver with 4 asynchronous threads. Agent is trained for 13.2M total steps, corresponding to 1 day of experience on a Quadro P5000 GPU device. 

There are three widely used types of goals for indoor navigation: PointGoal, ObjectGoal and RoomGoal. PointGoal instructs agent to go to a precise relative point, represented as $(x, y)$. ObjectGoal makes agent to find a object in house. RoomGoal, as the most complex task, requires agent to find a specified room house. RoomGoal is quite difficult not only because texture of furnitures are in diverse color but also difference of its shapes and positions can make distinct observations. Moreover, to navigate efficiently enough, agents have to learn how rooms are connected, such as a bedroom is always connected with a bathroom and a kitchen is often connected with a living room. As a result, we adopt RoomGoal as goal type to prove our idea. 

We test our baseline models on the sampled test episodes. Usually, performance of a navigation model is evaluated by success rate~\cite{anderson2018vision} report as percentage. 
Recent paper~\cite{anderson2018evaluation} proposes a new measurement called Success
weighted by (normalized inverse) Path Length(SPL). This metric requires model to navigate to goal as possible when taking the optimal path. We evaluate our models on both metrics. 
% Since we only have 18 rooms and 180 tests episodes, we propose a better way to evaluate the performance of our model. Since the action next step is sampled from the action distribution $a$, we test same model with 5 different random seeds. Then we calculate the means and the stand deviations of 5 results, which better describes the different predictions between models, makeing our results more convincing. 

\begin{figure}[t]
\centering
\includegraphics[width=0.98\linewidth]{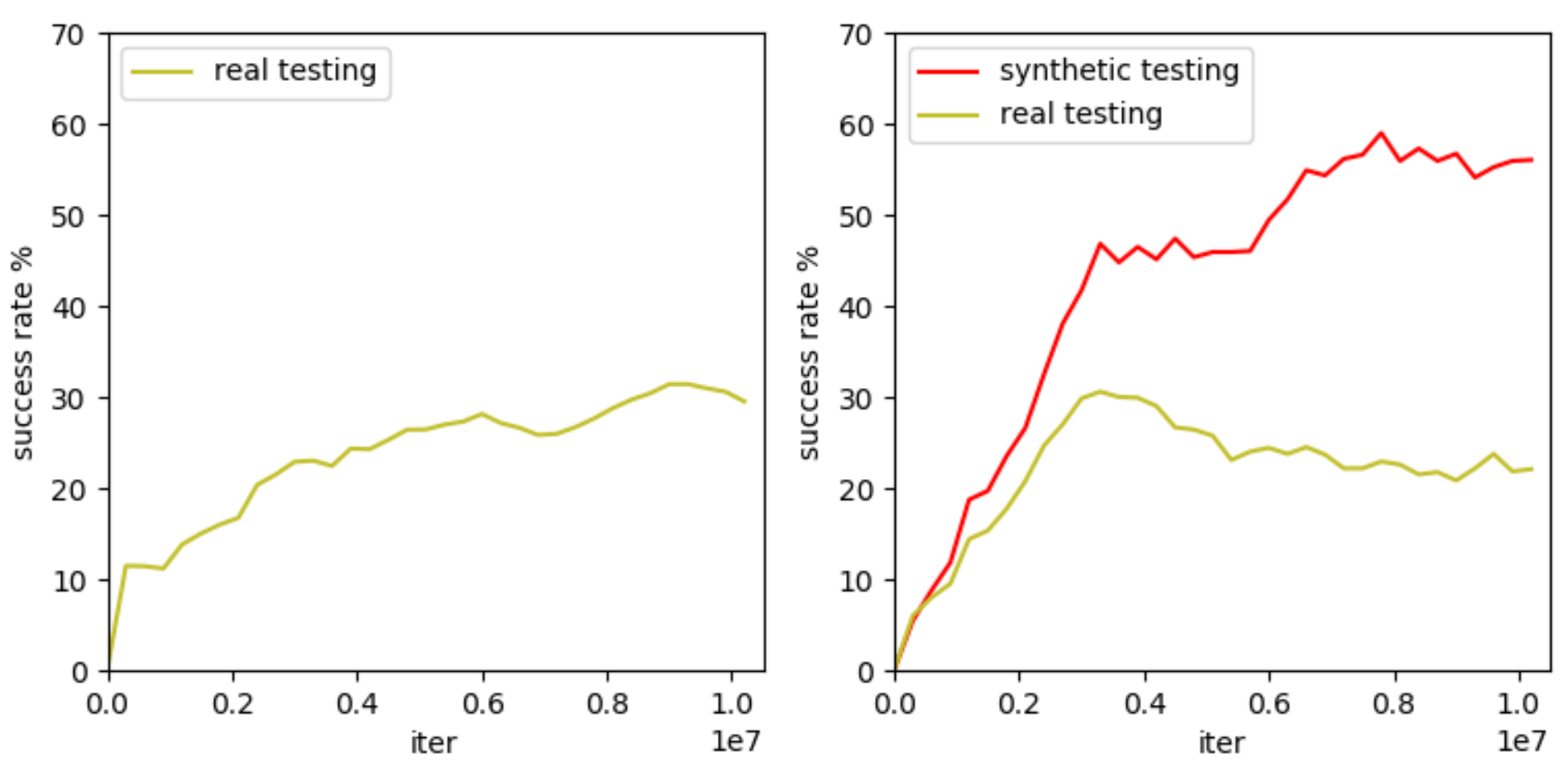}
    \caption{
 The left figure shows the testing results of real baseline on real environment. 
 The right figure shows the testing results of synthetic baseline on synthetic and real environment respectively. 
 %We test model for every 300K iterations. 
    }
\label{fig:success_rate}
\end{figure}

To show our baseline models are sufficiently trained, we test the success rate of our baselines for every 30K iterations. Figure~\ref{fig:success_rate} indicates both baselines converged after $8 \times 10^{6}$ iterations. However, synthetic baseline perform badly when tested on real environment and its performance continually drop after $8 \times 10^{6}$ iterations. This performance drop shows there is domain gap between synthetic and real environments. 

\subsection{Adversarial Feature Adaption Experiments}
For unpaired adversarial training, we sample 10,000 RGB images from synthetic and real environment respectively. 
We adopt Adam optimizer with initial learning rate = 0.0001, $\beta_{1}$ = 0.9,  $\beta_{2}$ = 0.999. Each training batch with size 64 consists of 32 synthetic images and 32 real images. This training process is easy to converge so that we test all the model at 1000 iterations. 

Note that we use initialize the parameter of $M_{r}$ with $M_{s}$ because $M_{s}$ content some semantic encoding. Without a good starting point, model is easily collapse and its performance could be even lower than baseline.

%\begin{table}[]
%\begin{tabular}{l|l|l}
%                                      & mean  & std  \\ \hline
%synthetic baseline                    & 24.13 & 2.12 \\ \hline
%real baseline                         & 32.67 & 0.73 \\ \hline
%synthetic baseline+fine-tune          & 36.8  & 2.64 \\ \hline
%synthetic baseline+feature adaptation & 56.27 & 2.94
%\label{table:adda}
%\end{tabular}
%\caption{Success rate for different baselines and feature %adaptation results}
%\end{table}

\begin{table}
\centering
\scalebox{1} {
\begin{tabular}{lcc}
\hline
\textbf{} method & \% success rate & \% SPL \\
\hline
sim baseline & 24.13\% $\pm$ 2.12\% & 22.46\% $\pm$ 1.52\% \\
real baseline & 32.67\% $\pm$ 0.73\% & 27.60\% $\pm$ 2.33\%\\
sim+FT & 36.80\% $\pm$ 2.64\% & 27.95\% $\pm$ 1.18\%\\
sim+FA  & {\bf 56.27\% $\pm$ 2.94\%} & 38.74\% $\pm$ 0.85\%\\
sim+PM  & 47.64\% $\pm$ 2.25\% & 33.15\% $\pm$ 2.37\%\\
sim+FA+PM  & {\bf 58.53\% $\pm$ 2.96\%} & {\bf 42.25\% $\pm$ 2.23\%}\\
\hline
\end{tabular} }
\caption {Success rate for different baselines and feature adaptation results. Here FT stands for finetune, FA means adversarial feature adaptation and PM means policy mimic. }
\label{table:adda}
\end{table}

\noindent\textbf{Comparison against baselines}
In Table~\ref{table:adda}, we compare three different baselines with our feature adaptation method. Among three baselines the best performance is achieved by fine-tuning on real environment after training on reinforcement synthetic environment. This baseline gets highest success rate because it is trained on both environment so that get more information. However, our method outperforms the best baseline greatly by 19.47\%, even though baseline with finetune is trained with more information. 

Note that baseline with fine-tune access more information than feature adaptation since fine-tune on real environment will not only get the image input but also get the reward supervision. 
Even though std is large since our scale of testing data is small, huge increasing of mean value showing feature adaptation method works on reinforcement policy transfer. 

\noindent\textbf{Analysis of the identity loss}
In Table~\ref{table:idt}, we compare the influence of different identity loss weight to adaption performance. We find that identity loss plays a important role in adversarial training. Without identity loss, which means its loss weight equals 0, performance will drop by 7.74\%. 

When identity weight goes even larger, mapping function for real environment $M_{r}$ tend to act much more like $M_{s}$. It makes feature distribution $F_{r}$ far away with $F_{s}$, which may led policy function $P_{r}$ make wrong decision. When identity weight continue increasing, Performance of the whole model will drop and eventually be equal to synthetic baseline. Thus we adopt the hyper-parameter identity loss weight = 0.0005 in the following experiments.

%\begin{table}[]
%\begin{center}
%\begin{tabular}{l|l|l}
%             & mean  & std  \\ \hline
%idt=0        & 48.53 & 2.93 \\ \hline
%idt=0.000005 & 54    & 2.23 \\ \hline
%idt=0.0005   & 56.27 & 2.94 \\ \hline
%idt=0.05     & 53.07 & 1.55 
%\end{tabular}
%\end{center}
%\caption{Ablation study: success rate for different weight for identity loss}
%\label{table:idt}
%\end{table}

\begin{table}
\centering
\scalebox{1} {
\begin{tabular}{lcc}
\hline
\textbf{} idt ablation & \% success rate \\
\hline
idt=$0$ & 48.53\% $\pm$ 2.93\% \\
idt=$5*e^{-6}$ & 54.00\% $\pm$ 2.23\%  \\
idt=$5*e^{-4}$ & {\bf 56.27\% $\pm$ 3.94\%} \\
idt=$5*e^{-2}$  & 53.07\% $\pm$ 1.55\%\\
\hline
\end{tabular} }
\caption {Ablation study: success rate for different weight of identity loss}
\label{table:idt}
\end{table}

\subsection{Policy Mimic Experiments}
Following the Policy Mimic training described in the Section \ref{mimic}, we only update policy function $P_{r}$ in real environment.  

The environmental configuration and model hyper-parameter is same as baseline when trained in real environment. We use $M_{r}$ trained in Adversarial Feature Adaptation method and initialize $P_{r}$ by $P_{s}$. Similar to feature adaptation, if we just random initialize our policy function, the result could be much lower. To prove our policy mimic method is general, we test our method on both A3C model and UNREAL model.

%\begin{table}[]
%\begin{center}
%\begin{tabular}{l|l|l}
%                      & mean  & std  \\ \hline
%UNREAL baseline       & 24.13 & 2.12 \\ \hline
%UNREAL baseline+FA    & 56.27 & 2.94 \\ \hline
%UNREAL baseline+FA+PM & 58.53 & 2.96 \\ \hline
%A3C baseline          & 16.00    & 2.02 \\ \hline
%A3C baseline+FA       & 34.80  & 0.88 \\ \hline
%A3C baseline+FA+PM    & 50.00    & 1.88
%\end{tabular}
%\end{center}
%\caption{Success rate for different baselines, feature %adaptation(FA) and policy mimic(PM) results}
%\label{table:pm}
%\end{table}

\begin{table}
\centering
\scalebox{1} {
\begin{tabular}{lcc}
\hline
\textbf{} method & \% success rate \\
\hline
UNREAL baseline & 24.13\% $\pm$ 2.12\% \\
UNREAL baseline+FA & 56.27\% $\pm$ 2.94\%  \\
UNREAL baseline+FA+PM & {\bf 58.50\% $\pm$ 2.96\%} \\
A3C baseline  & 16.00\% $\pm$ 2.02\% \\
A3C baseline+FA  & 34.80\% $\pm$ 0.88\% \\
A3C baseline+FA+PM  & {\bf 50.00\% $\pm$ 1.88\%} \\
\hline
\end{tabular} }
\caption {Success rate for different baselines, feature adaptation(FA) and policy mimic(PM) results}
\label{table:pm}
\end{table}

\noindent\textbf{Comparison against baselines}
Table~\ref{table:pm} summarizes our results and compares our policy mimic method with feature adaptation only and baseline. 

UNREAL baseline is higher than A3C baseline for 8\% since the auxiliary task provides additional training signals. 
However, after feature adaptation, the performance gap is increased to nearly 22\%. A reasonable explanation is additional training signals are mainly beneficial to policy network. Thus feature transfer can be a greater help for UNREAL model compared to A3C model. 

Finally, it is the policy mimic method that reduces the performance gap. Even though the teacher model for policy mimic of A3C is A3C baseline trained on synthetic environment, model can still absorb useful knowledge learned from synthetic data.

%\begin{table}[t]
%\begin{center}
%\begin{tabular}{l|l|l}
%                   & mean  & std  \\ \hline
%UNREAL baseline    & 24.13 & 2.12 \\ \hline
%UNREAL baseline+FT & 36.8  & 2.64 \\ \hline
%mimic\ weight=0    & 36    & 3.55 \\ \hline
%mimic\ weight=0.1  & 58.53 & 2.96 \\ \hline
%mimic\ weight=0.2  & 34.53 & 1.95 \\ \hline
%mimic\ weight=0.5  & 22.8  & 1.85
%\end{tabular}
%\end{center}
%\caption{Ablation study: success rate for different weight for %mimic loss}
%\label{table:mimcloss}
%\end{table}

\begin{table}
\centering
\scalebox{1} {
\begin{tabular}{lcc}
\hline
\textbf{} mimic ablation & \% success rate \\
\hline
UNREAL baseline & 24.13\% $\pm$ 2.12\% \\
UNREAL baseline+FT & 36.80\% $\pm$ 2.64\%  \\
mimic\ weight=0 & 36.00\% $\pm$ 3.55\% \\
mimic\ weight=0.1  & {\bf58.53\% $\pm$ 2.96\%} \\
mimic\ weight=0.2  & 34.53\% $\pm$ 1.95\% \\
mimic\ weight=0.5  & 22.80\% $\pm$ 1.85\% \\
\hline
\end{tabular} }
\caption {Ablation study: success rate for different weight for mimic loss. Here FT means baseline with finetune. }
\label{table:mimcloss}
\end{table}

\noindent\textbf{Analysis of the mimic loss}
Table~\ref{table:mimcloss} reports our ablation experiment upon mimic loss. Compared with ablation experiment for weight of identity loss, weight of mimic loss seems to be more sensitive. Performance will drop rapidly if mimic weight be slight deviate the optimal value. When mimic loss weight is 0, it is equivalent to finetuning model on real environment. Compared to finetune baseline above, since mapping function  $M_{r}$ has been transferred, the performance is slightly higher. 

The reason why mimic loss is rather important is that the policy function can overfit easily in the real environmental without rich training data.  Thus imitating behavior of model trained on synthetic model can solve this difficulty. 

\subsection{Visualization}

\begin{figure*}[t]
\centering
\includegraphics[width=0.98\linewidth]{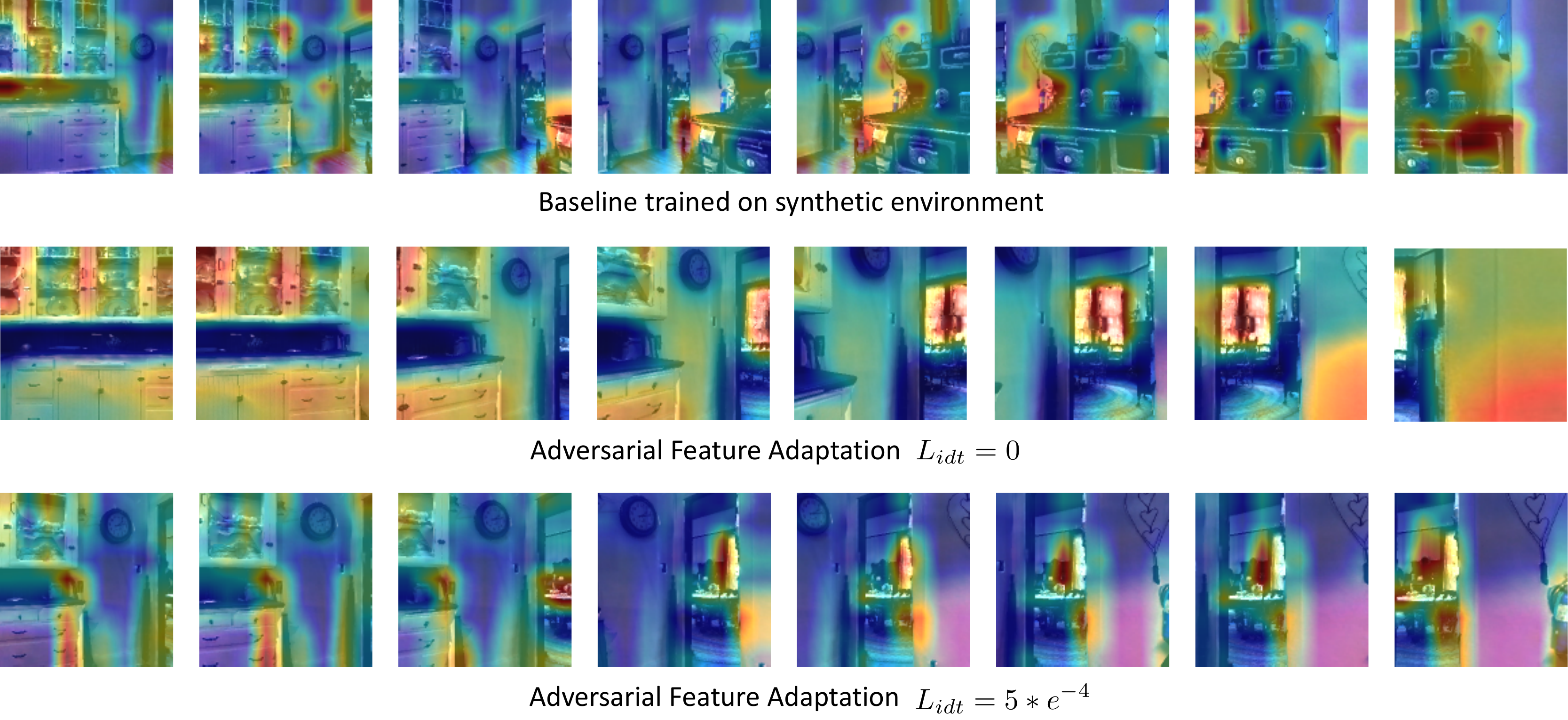}
    \caption{
Heatmap for the last convolution in mapping function. Red and yellow regions are places with large values where network mainly focus on. We can capture the relationship between scene transitions from left to right and attention on heatmaps. 
    }
\label{fig:vis_1}
\end{figure*}

\begin{figure*}[t]
\centering
\includegraphics[width=0.98\linewidth]{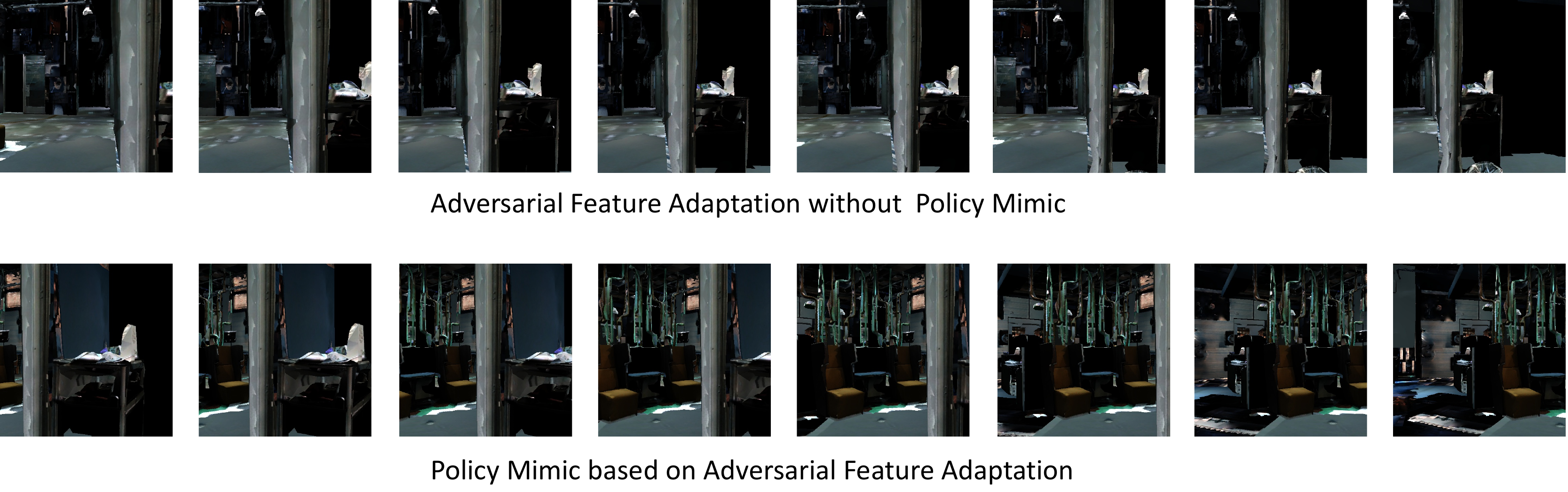}
    \caption{
RGB image sequences from left to right indicates the models' trajectories in testing. 
    }
\label{fig:vis_2}
\end{figure*}

In this section, we evaluate our methods from qualitative aspects. 

We run different models on unseen real environments in the test set. Meanwhile, we record agents navigating trajectories, RGB input sequences and some intermediate results. By comparing these results, we testify that our methods are able to optimize intermediate procedure and therefore, improve final performance. 

\noindent\textbf{Visual Attention}
We now attempt to analysis how adversarial feature adaptation influence mapping function. 

We plot sequences of heatmap for last convolution layer combining with RGB inputs. Figure~\ref{fig:vis_1} compares three different models with similar RGB images input. 

It is obvious that baseline trained on synthetic data is not able to embedding real environment data. The model can not focus on specific targets. On the contrary, its attention is scattered on the whole map. Finally, it can not go to target room but miss the door by keep turning right.

After adversarial training even though without identity loss, model learns to focus on some targets like cabinet, wall and window. However, these targets have minor effect on navigation. It means model can learn how to distinguish objects by adversarial adaptation. By paying too much attention on the wall at right, this model misses to cross the door either. 

With identity loss augmenting adversarial training, model not only learns to recognize objects, but also know the semantic difference between targets. We find that this model pay more attention on door like edges and be able to find the real door.  

\noindent\textbf{Policy Behavior}
Based on model after adversarial adaption, we now analysis the policy behavior in some complex cases. Figure~\ref{fig:vis_2} shows a scene where model need to cross a door not being able to distinguish from RGB information easily. 

Model without policy mimic does not have a stable navigation behavior. It repeatedly turns left or right in a random way. It seems that model is overfit on a local minimum, which prohibit its exploration. 

In contrast, model with policy mimic gains a more stable navigation. It keeps turning following one direction to explore scene of whole room.

This knowledge is learned from large scale synthetic data, where model is able to be trained among diverse scenes to avoid overfitting. The knowledge is suitable for navigation in real environment, however, which model can hardly learn from because of few training data.

\section{Conclusion}

%It the synthetic data is much easier and cheaper than the real data to obtain, models only use real trained in real environment are easy to overfit. 

To relieve the problem of lacking real training data, we propose JRT to discover the knowledge learned from synthetic reinforcement environment to facilitate the training process in real environment. Specifically, we integrate adversarial feature adaptation and policy network into a joint network. 
We demonstrated the effectiveness of the framework with extensive experiments. We also conducted ablation studies of the identity loss and mimic loss to show its superiority.
Finally, our methods outperform baselines in both qualitative and quantitative ways without any additional human annotations. 

In the future, we will validate the transfer ability on other tasks, \eg, Embodied Question Answering~\cite{das2018embodied} and exploit temporal information for feature adaptation~\cite{wang2018video, zhu2017uncovering, chang2017semantic} and feature analysis~\cite{chang2017semisupervised}.

Also, policy mimic can be further improved by considering episode-wise optimization. In addition, it is meaningful to investigate how to transfer policy in vision and language navigation tasks.

\noindent\textbf{Acknowledgments.}
We thank AWS Cloud Credits for Research for partly supporting this research.

%\small{\noindent{\textbf{Acknowledgments}}.
%We thank AWS Cloud Credits for Research for partly %supporting this research.
%}

%(1) MP3D from scratch.
%Details
%(2) SUNCG simple transfer. Details.
%(3) ADDA
%- Figure 1 (A3C) (curve)
%- Figure 2 (UNREAL)
%- Analysis
%\subsection{Adaption+Policy estimation Results}
%(1) Pick a room setting. Show the ADDA+Policy results. Compare with baseline.
%A3C, UNREAL. (Table)

\end{document}